# CottonLeafVision: An Explainable and Robust Deep Learning Framework for Cotton Leaf Disease Classification

Rafi Ahamed
*Dept. of CSE*
*East West University*
Dhaka, Bangladesh
fidaahamed15@gmail.com

Md. Abir Rahman
*Dept. of CSE*
*East West University*
Dhaka, Bangladesh
abirrahman.sakin@gmail.com

Tasnia Tarannum Roza
*Dept. of CSE*
*East West University*
Dhaka, Bangladesh
tasniatarannumroza28@gmail.com

Munaia Jannat Easha
*Dept. of CSE*
*East West University*
Dhaka, Bangladesh
jannateasha004@gmail.com

Md. Asif Khan
*Dept. of CSE*
*East West University*
Dhaka, Bangladesh
ak92452466@gmail.com

Sudeepta Mandal
*Dept. of CSE*
*East West University*
Dhaka, Bangladesh
msudeep996@gmail.com

***Abstract*— Globally, cotton is a highly economically beneficial crop, as the textile industry heavily depends on it. So, the precise identification and detection of cotton leaf disease is crucial for economic stability. The development goal of "CottonLeafVision" is to accurately classify and detect cotton leaf disease. With this goal, we have evaluated multiple pretrained Deep Convolutional Neural Networks, including DenseNet201, InceptionV3, and VGG19 on a publicly available cotton leaf disease image dataset. This image dataset includes seven classes, six disease classes, and one healthy class, collected under various field conditions reflecting real-world challenges. Among these pretrained models, with DenseNet201, we have achieved the highest classification accuracy of 98%. To enhance the model reliability and interpretability, we have implemented different techniques and methods such as Gradient-weighted Class Activation Mapping (Grad-CAM), occlusion sensitivity analysis and adversarial training to increase the noise resistance of the model. Finally, we have developed a prototype in order to utilize the model's capabilities on real life agriculture. This paper shows the deep learning model's capabilities to classify the disease in real-life cotton disease management situations.**

## I. INTRODUCTION

Cotton (Gossypium spp.), one of the most important commercial crops in the world, is a vital source of edible oil and animal feed, as well as the primary natural fiber for the global textile industry. However, several diseases are becoming a greater threat to the fiber quality and productivity of this important crop worldwide. Because leaves show signals of more than 80-90% of cotton plant diseases, researchers believe that leaves are the most important markers of plant health. Traditional plant disease diagnosis continues to rely heavily on professionals who visually and physically investigate plants. Certain approaches are expensive, time-consuming, and require a significant amount of manual labor. Large agricultural areas creates significantly more challenges. For that reason, farmers detect illness symptoms when it has progressed significantly. It makes treatment more complicated, expensive, and likely to result in significant decreases in overall crop output.

Emerging challenges in cotton farming have increased the need for intelligent, high-speed diagnostic tools capable of identifying plant diseases with precision. Modern computer-vision and machine-learning techniques make it possible to automatically analyze images of affected cotton leaves, recognizing various disease categories and evaluating their severity. The availability of reliable, expertly annotated image datasets is critical to the performance of these systems, as it allows the algorithms to understand minor visual distinctions between circumstances. This plan suggests establishing an administrative diagnosis platform monitored by experienced professionals. Farmers can make an immediate assessment and get expert recommendations by uploading the image. Initially, crop experts and computer scientists will collaborate to share knowledge about disease detection and prepare reliable training data. These smart systems will act as decision support platforms, which help farmers to identify diseases early and receive effective advice for their management. These type of systems intends to establish exact models that can perfectly detect cotton ailments in their early phases. This type of method can increase farmers' crop productivity, reduce disease-based losses, and increase the production of cotton that is favorable to the environment. It ensures that people globally have access to enough food and fiber. In our paper, we raised two main questions: (1) Is it possible to remain optimized to gain higher accuracy when we test this on real world? (2) What will be the performance of our model compared to other CNN models in terms of accuracy while detecting cotton leaf diseases?

## II. RELATED WORKS

The SAR-CLD-2024 dataset, which is introduced by Bishshash et al., is a large, field-based collection of cotton leaf images. It is covered with multiple disease types and healthy leaves. They raised accuracy around 96% with strong scores [1].

Abudukelimu et al. proposed CMYOLO, an improved YOLOv8 model trained to detect small and subtle cotton leaf disease spots more accurately and efficiently. Their model CMYOLO reaches about 0.933 mAP50 with high precision and recall, which outperforms several popular detectors while remaining fast enough for real-time use in the field [2]. A fine-tuned VGG16-based deep learning model introduced by Kaur et al. for arranging cotton leaf images into three disease classes, which are: bacterial blight, curl virus, Fusarium wilt, and the rest of all are healthy leaves. They trained a Kaggle dataset, which contains 1,711 images. On this dataset, they applied two augmentation operations per image: rotation and reflection. They expanded it to about 5,000 images to improve robustness. Their model achieved about 95.5% accuracy [3].

They highlighted that the fine-tuned VGG16 model can distinguish healthy from diseased cotton leaves. Herok and Ahmed proposed a CNN model that is trained on a dataset with seven disease classes plus healthy leaves. They worked on a challenging dataset of 6,158 images (8 classes), which is collected from heterogeneous online sources. Several pretrained architectures (InceptionV3, ResNet152V2, InceptionResNetV2, Xception, MobileNetV2, DenseNet121, VGG16) were compared. Among them, VGG16 achieved the best test accuracy of about 95.02% [4]. They stated that this transfer learning pipeline with the curated dataset offers an accurate solution for early cotton leaf disease detection. It can also be extended to other plant disease tasks in future updates.

Kaur et al. developed a VGG16 deep learning model to classify four major cotton leaf diseases: bacterial blight, Verticillium wilt, Fusarium wilt, and cotton leaf curl disease automatically. They preprocessed a large dataset of cotton leaf images, which are labeled. After that, they applied augmentation. The model achieved about 93.8% overall accuracy, with precision, recall, and F1-scores all above 92% [5]. They argue that the finetuned VGG16 model can serve as a practical tool for early, automated diagnosis of key cotton diseases. It can also support precision agriculture by reducing losses.

Manjula Devi et al. developed a deep learning–based CNN model, which is trained on a labeled dataset of healthy and diseased leaves. It was collected under varying conditions. After applying preprocessing, augmentation, and transfer learning, we compared configurations to improve robustness and accuracy. Finally, their model reached about 97% accuracy [6]. Walia et al. introduced a VGG16- based deep learning architecture that classifies four major cotton leaf diseases: leaf curl, bacterial blight, leaf spot, and wilt. They trained and preprocessed a 1,700-image dataset.

For this specific multi-class task, they applied augmentation and fine-tuned VGG16 using transfer learning. The final model gained around 89–90% overall accuracy [7]. They assured that their VGG16 architecture offers a reliable tool for early cotton disease detection. Mehmood et al. carried out a comparative study of five hand-crafted feature extraction methods using an SVM classifier. They collected a 2,400-image dataset from Kaggle and applied augmentation to fix class imbalance.

Their experiments show that Gabor wavelet–based features give the best performance. It reaches about 92% accuracy[8] among all methods. They recommend Gabor wavelet features plus SVM as an effective, low-cost approach for automatic cotton leaf disease identification.

## III. Methodology

Fig 1 shows the whole work process of the proposed classification architecture. For finding out the cotton leaf disease using a publicly available dataset, this study utilizes a deep learning-based approach. We preprocessed the images by normalizing and resizing to increase the model's robustness. We split the data into training, validation, and test sets. Then we balanced the training data classes through the oversampling method.

After that, we used three pretrained models, DenseNet201, InceptionV3, and VGG19. We trained these models with distinct hyperparameters and selected the best model based on the validation metrics. Adversarial training was also applied to improve resistance to noise and confusion. Finally, we used Grad-CAM to generate visual explanations by using Grad-CAM and highlight the image regions that influence the model's predictions and enhance the precision.

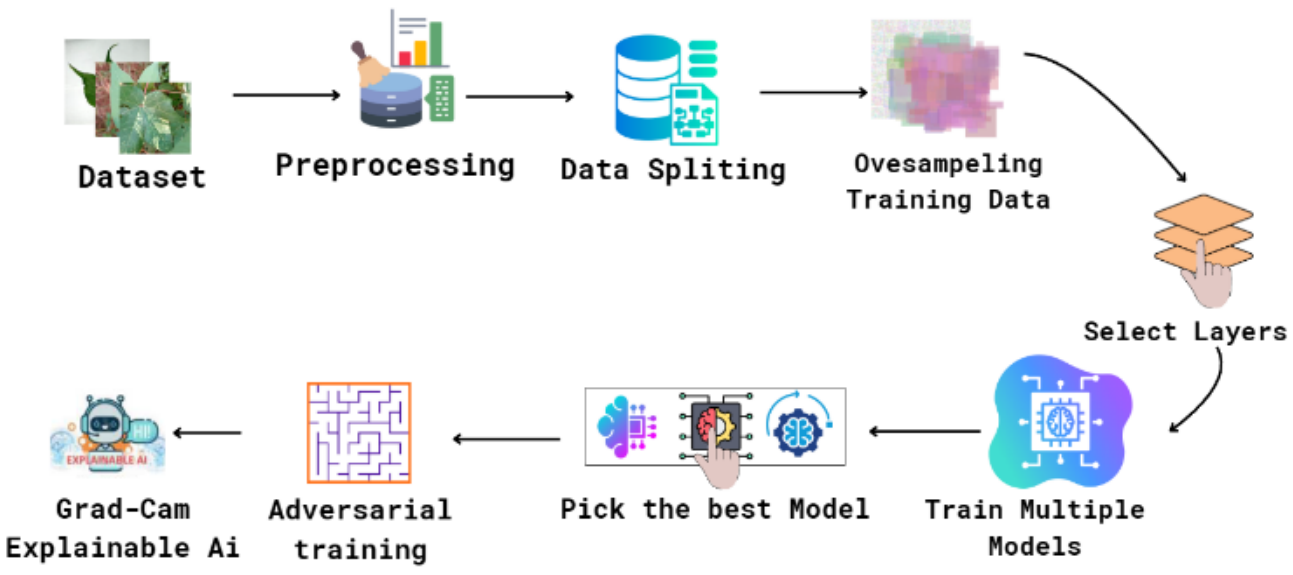

Fig. 1. Methodological framework

### A. *Dataset*

In this research, we used an augmented dataset publicly available for Cotton Leaf Disease Detection, shown in Fig. 2. In its original dataset, there are 2137 images divided into seven classes, covering six cotton leaf diseases and healthy leaves[9].

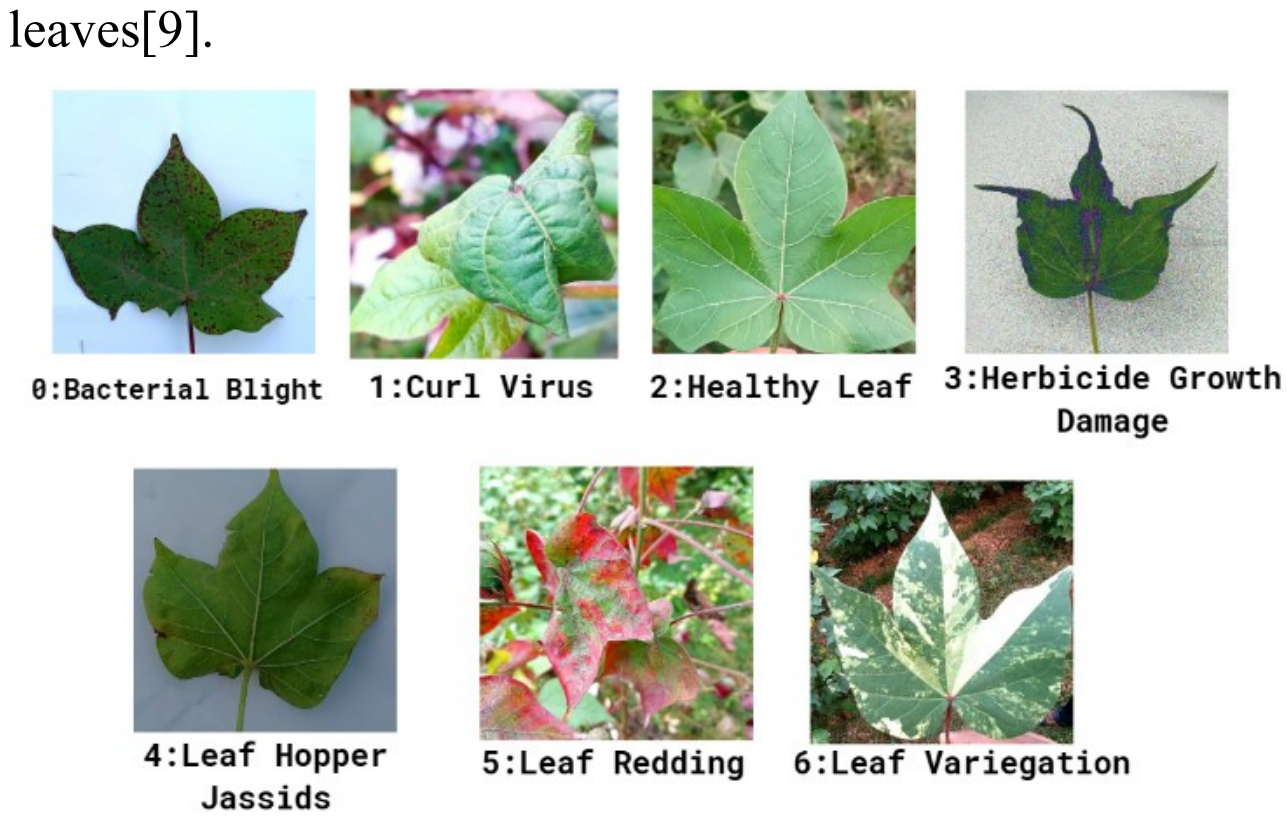

Fig. 2. Sample images from the dataset

The augmented dataset, it has 7000 augmented images, enhancing the effectiveness of deep learning models for precise classification and diagnosis of cotton leaf diseases. All the images were resized to 224x224 pixels for training models, and the pixel values were normalized to enhance training stability. After that, we normalized the pixel values to enhance training stability. The dataset was also divided into three parts: training, testing, and validation. This dataset has set a benchmark for developing accurate and generalized models for Cotton leaf disease and classification.

### B. *Preprocessing*

To match the input requirements of the pre-trained models, all photos were downsized to 224x224 pixels. To enable stable optimization, they were also normalized to the range of [0, 1]. To ensure that the evaluation was carried out on previously unseen samples, the dataset was split into subsets consisting of 70% training, 20% validation, and 10% testing. By this processing, we have reduced the effects of differences in image contrast and resolution. We have also ensured evenness across all architectures.

### C. *Model Architectures*

We have assessed three well-known CNN architectures in order to evaluate the models' performances.

- DenseNet201 is a 201-layer deep convolutional neural network with dense connectivity that reduces vanishing gradients and encourages feature reuse [10].
- The VGG19 architecture is a deep convolutional neural network (CNN) with 19 weight layers, including 3 fully connected layers and 16 convolutional layers with tiny 3x3 filters [11]. It is intended for large-scale image categorization and usually takes 224x224 pixel RGB pictures as input.
- InceptionV3 is a 48-layer deep convolutional neural network that effectively achieves high image classification accuracy by using modular "Inception modules" with factorized and asymmetric convolutions, dimensionality reduction strategies, and auxiliary classifiers [12].

These architectures were selected for their proven success in leaf disease imaging tasks and complementary design characteristics.

### D. *Training Setup*

The same training method was used for all three models. The Adam optimizer has a learning rate of 0.0001. Categorical cross-entropy is a loss function suitable for tasks with multiple classes. The batch size was set to 32. To prevent overfitting, early stopping was used with patience for ten epochs. The number of epochs was 50 for all models, as shown in Table I.

TABLE I. HYPERPARAMETER SETTINGS FOR THE EVALUATED CNN ARCHITECTURES

| Parameter | VGG19 | InceptionV3 | DenseNet201 |
|---|---|---|---|
| Batch size | 32 | 32 | 32 |
| Loss function | Categorical cross-entropy | Categorical cross-entropy | Categorical cross-entropy |
| Learning rate | 0.0001 | 0.0001 | 0.0001 |
| Optimizer | Adam | Adam | Adam |
| Number of epochs | 50 | 50 | 50 |
| Early stopping (patience) | 10 | 10 | 10 |

## IV. RESULTS

This section evaluates the performance of the three CNN models used in this study. In addition, it provides an analysis of the strength of the most effective model

### A. *VGG19*

Fig. 3 illustrates the training and validation performance of VGG19. The loss and accuracy curves show stable convergence, with the validation accuracy stabilizing between 90% to 94%. This indicates the model could be effectively generalized without significant overfitting.

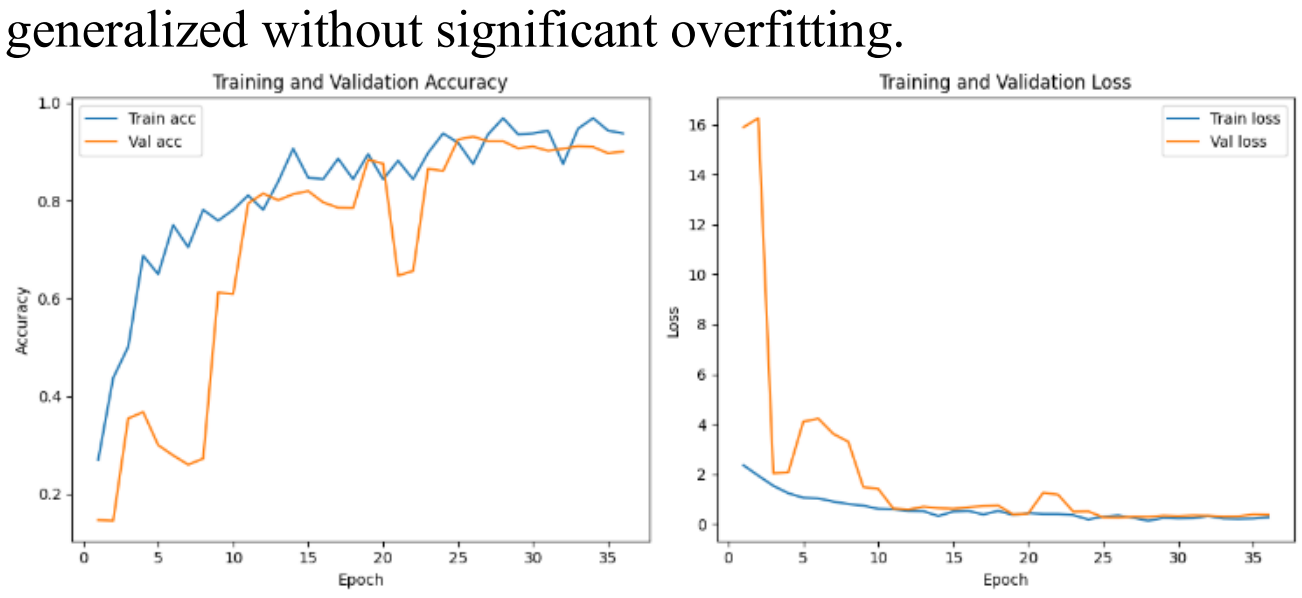

Fig. 3. Loss for training and validation of the VGG19 architecture

The Confusion matrix presented in Fig. 4 reveals a moderate accuracy across most classes, although some significant misclassification can be observed in The Leaf Redding and Leaf Hopper Jassids.

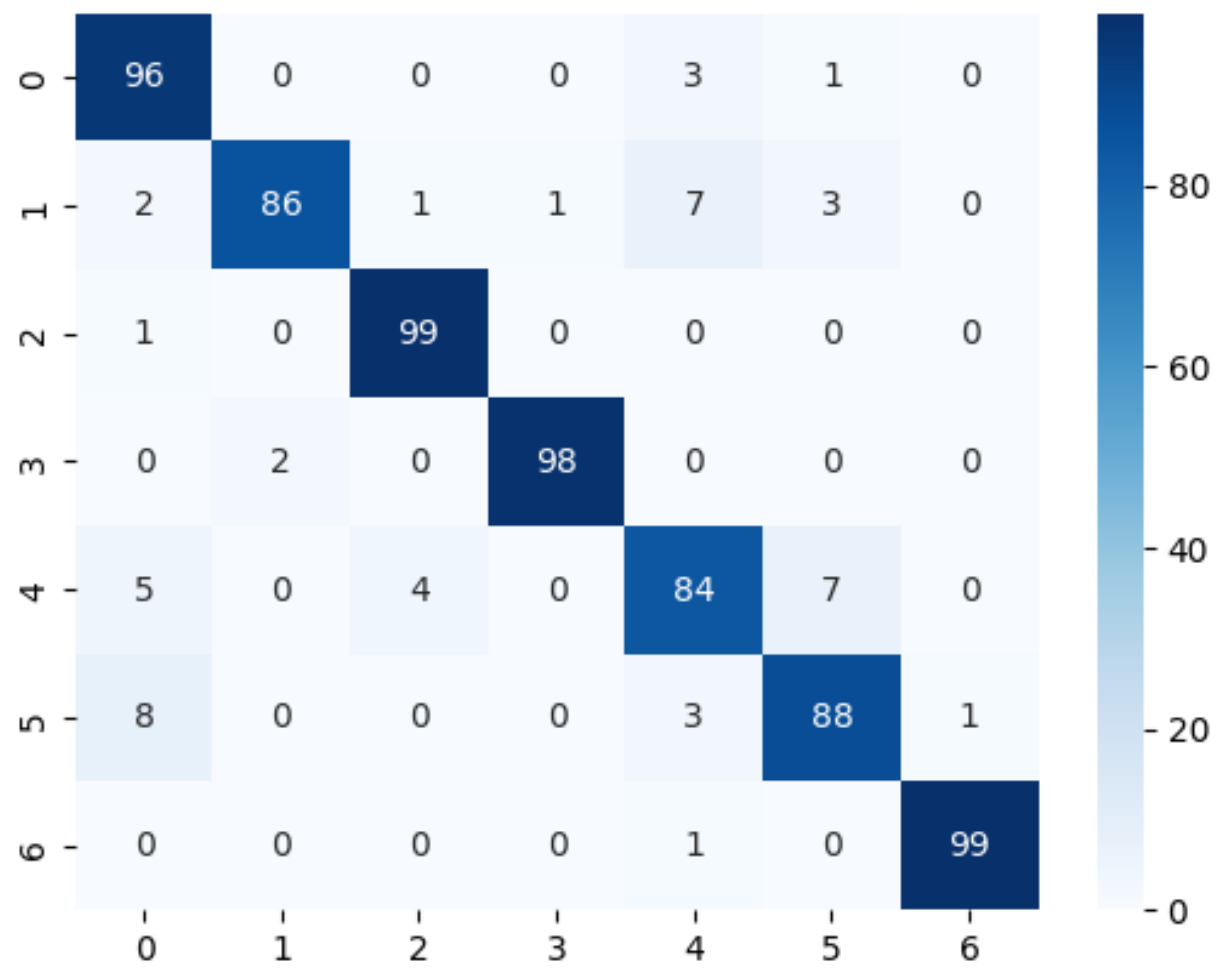

Fig. 4. Confusion Matrix of VGG19 architecture

### B. *InceptionV3*

In Fig. 5, InceptionV3 demonstrates strong convergence for training and validation accuracy, whereas validation accuracy consistently exceeds 95% and stabilizes around 97%

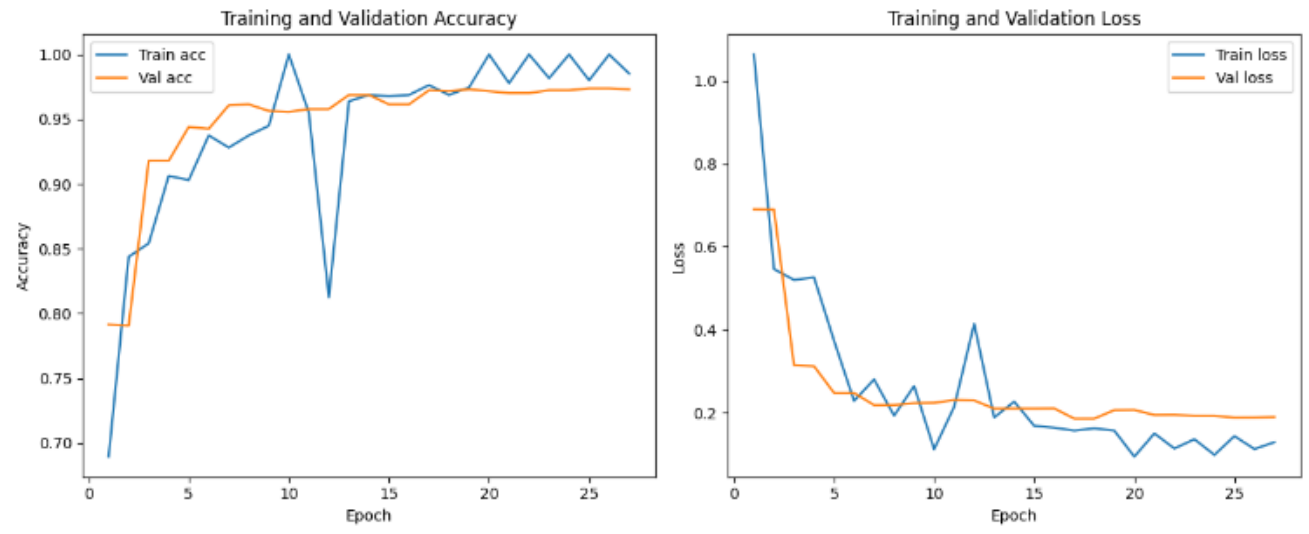

Fig. 5. Loss for training and validation of the InceptionV3 architecture

In Fig. 6, the confusion matrix indicates a balanced recognition across all diseases. InceptionV3 exhibited significantly superior performance in classifying disease, in comparison to VGG19, although minor overlaps can also be observed.

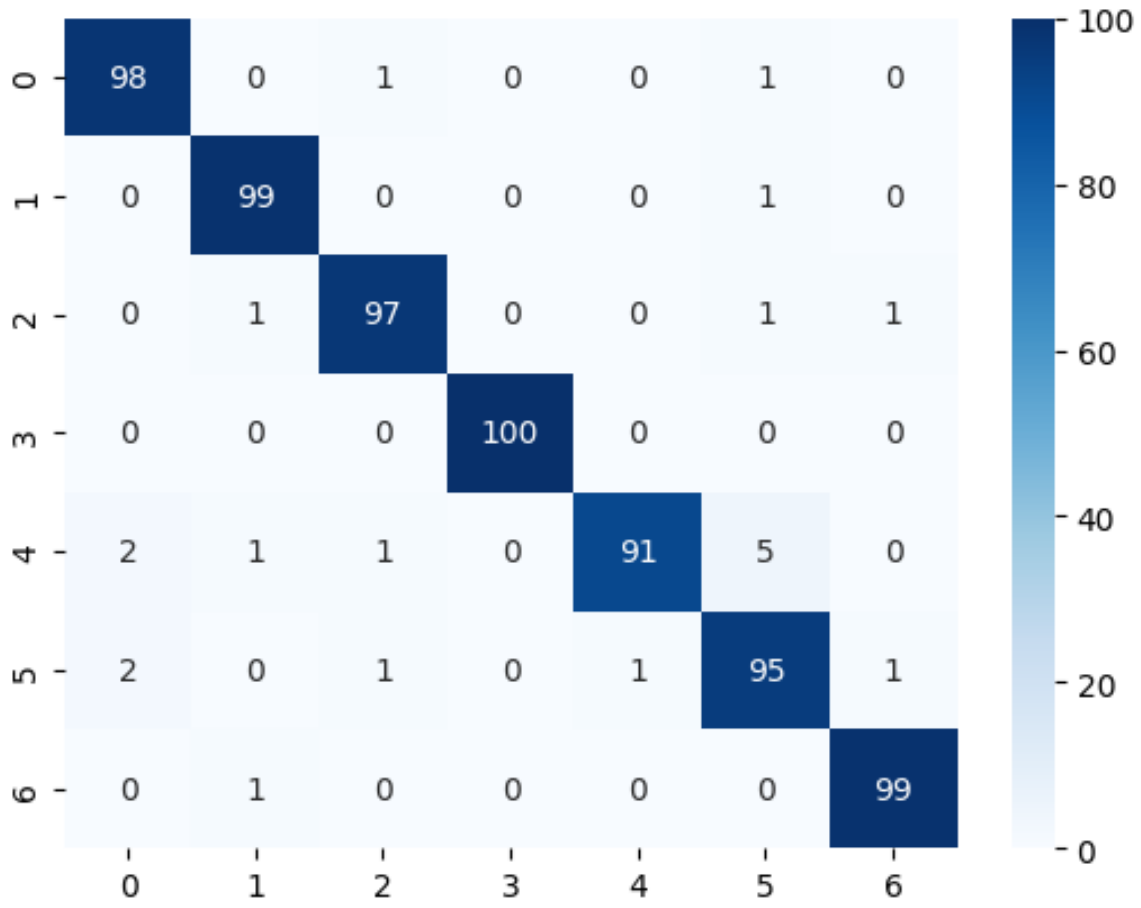

Fig. 6. Confusion Matrix of InceptionV3

### C. *DenseNet201*

DenseNet201 demonstrated the best overall performance among the evaluated models. As shown in Figure 7, the training and validation curves converged quickly. Here, the precision exceeds 97%, and the loss stabilizes at minimal values. The overall accuracy of the model stabilizes at 98%.

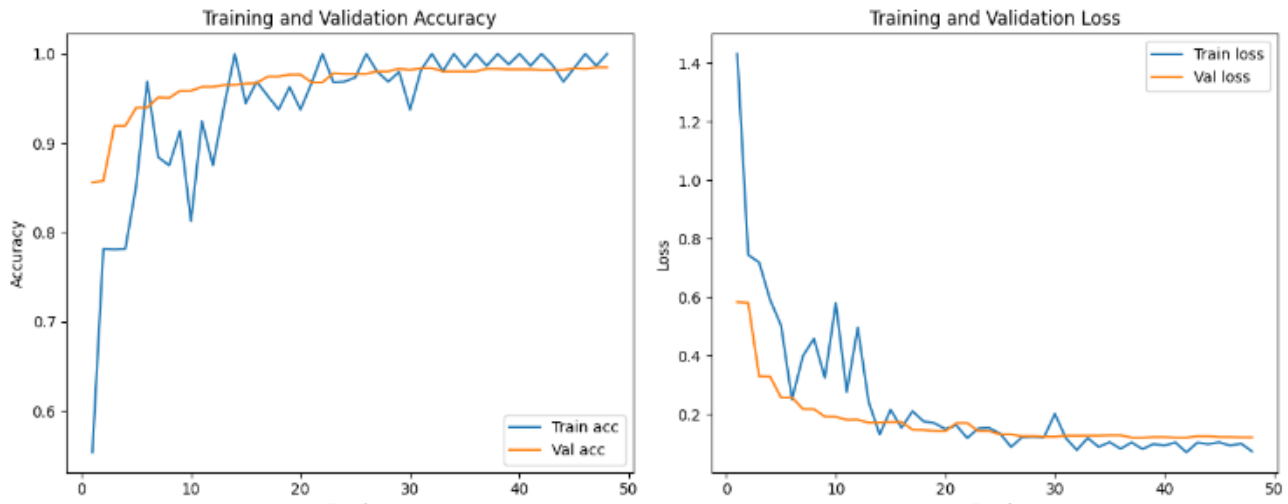

Fig. 7. Training and validation accuracy and loss curve of DenseNet201

We can observe excellent recognition of the disease in the confusion matrix, Fig. 8.

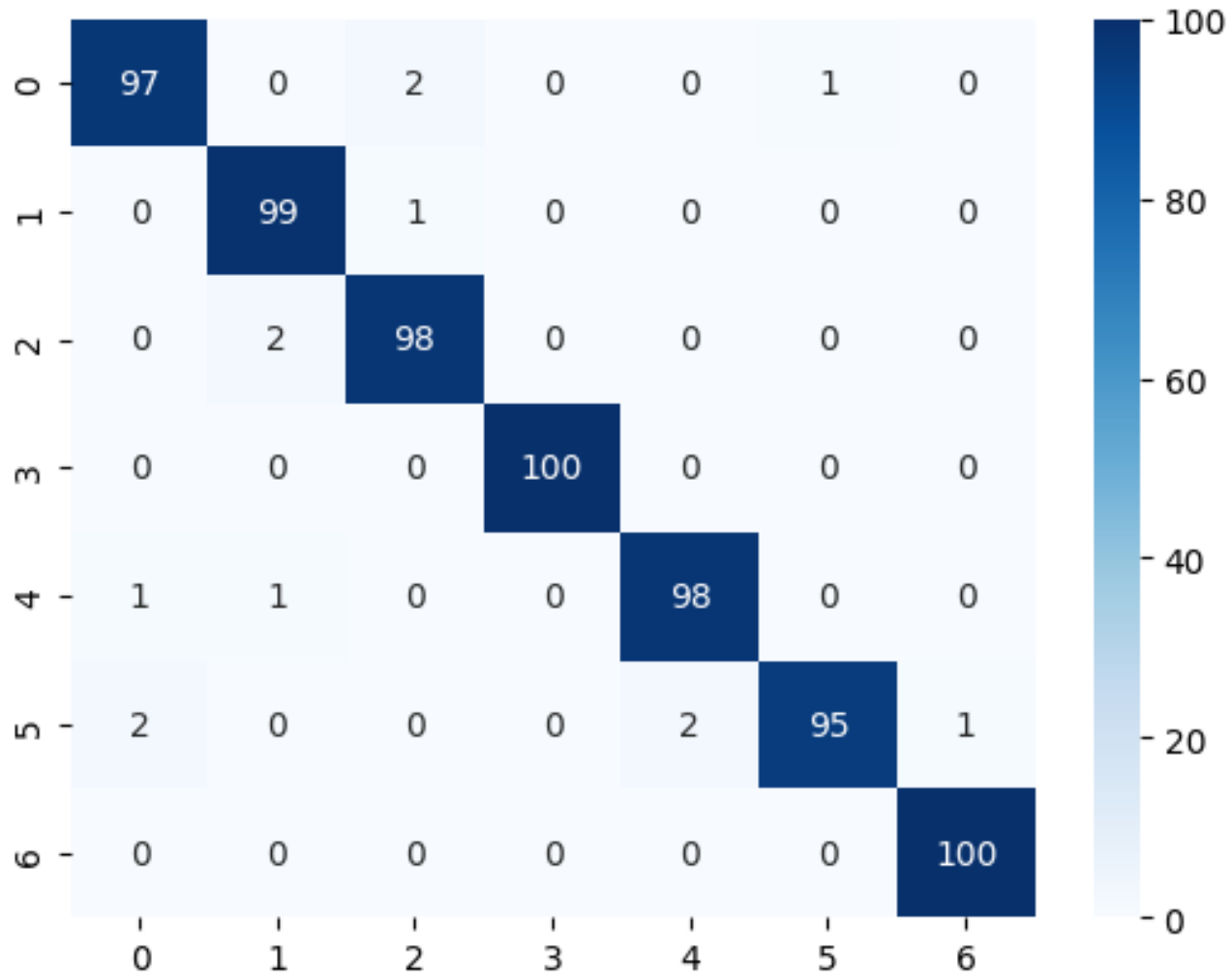

Fig. 8. Confusion Matrix of DenseNet201

The higher accuracy and balanced classification performance of DenseNet201 establish it as the most reliable model for Cotton Leaf Disease detection.

### *D. Adversarial training*

An adversarial technique was applied to DenseNet201 to further test reliability, and the results are shown in Table II. We can observe a consistent high validation accuracy across all ε values, and the model has achieved a peak accuracy of 98.91% at ε= 0.1. Even at high perturbation levels, the performance remained above 98.8%, and the loss is also very minor. These findings confirm that DenseNet201 is not only the most accurate but also resilient to input noise, thereby reinforcing its suitability for agricultural support.

TABLE II. ADVERSARIAL ROBUSTNESS EVALUATION OF DENSENET201 UNDER VARYING PERTURBATION LEVELS ( $\varepsilon$)

| Epsilon Value (ε) | Validation Loss | Validation Accuracy | Optimal Epochs |
|---|---|---|---|
| 0 | 0.1223 | 0.9833 | 21 |
| 0.1 | 0.1171 | 0.9891 | 34 |
| 0.12 | 0.1078 | 0.9898 | 44 |
| 0.14 | 0.1072 | 0.9884 | 49 |
| 0.16 | 0.1096 | 0.9898 | 31 |
| 0.18 | 0.1055 | 0.9906 | 23 |
| 0.2 | 0.1073 | 0.9906 | 3 |

## V. EXPLAINABLE AI

In addition to being accurate and reliable, models must be easy to understand for use in agriculture and farming applications. To render the deep learning models less like a "black box," we used a method called Grad-CAM[13]. This helps us determine which parts of the cotton leaf affect the model's predictions the most.

### *A. Grad-CAM Visualizations*

Fig. 9 shows Grad-CAM overlays on the cotton leaf. The heat map highlights the infected and affected regions of the leaf that influence the prediction of the model. In general, the highlighted areas are spread out showing no diseases, but here it highlights the affected area, which means the model is processing the images properly and avoiding the unnecessary background.

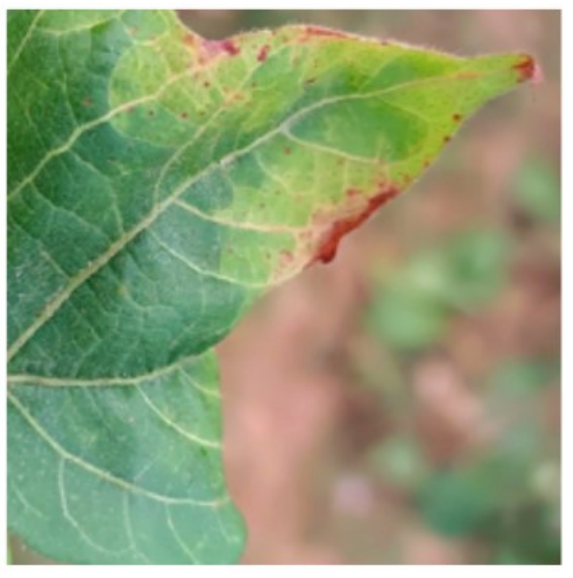
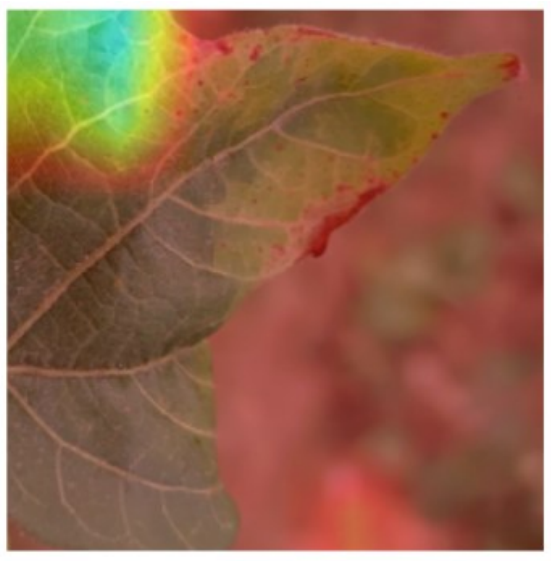

Original image Grad-CAM output

Fig. 9. Cotton Leaf Disease identification using Grad-CAM

### *B. Occlusion Sensitivity*

To demonstrate that our model is easy to understand, we used occlusion sensitivity analysis shown in Fig. 10. This means that we covered parts of the input leaf image to determine how it affected the confidence of the model in its predictions. As shown in Fig.10, the accuracy of the model decreased when important areas such as necrotic spots, fungal patches, insect feeding marks, and discoloration were covered. This shows that the model makes decisions based on the important parts of the leaf image.

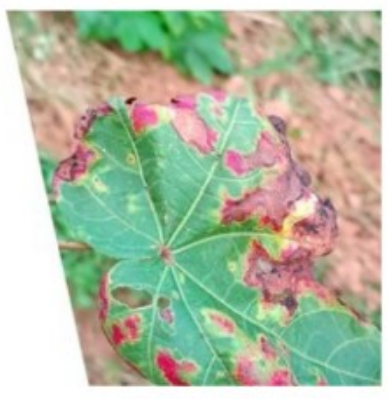
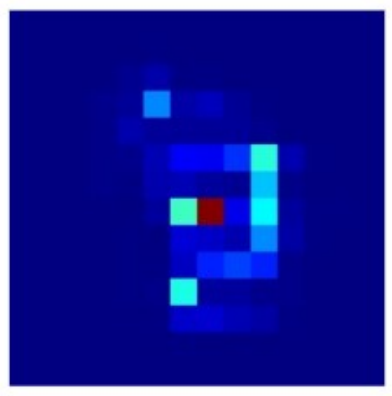
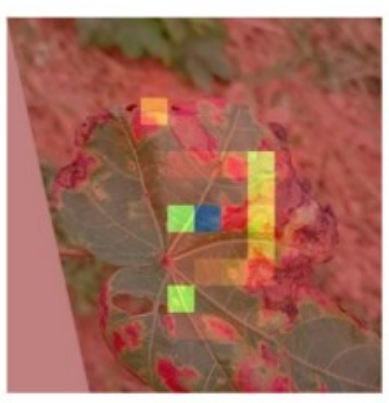

Original Occlusion Heatmap Occlusion Overlay

Fig. 10. Cotton Leaf Disease Identification Using Occlusion Sensitivity

Grad-CAM and occlusion sensitivity ensure that DenseNet201's predictions were accurate and easy for farmers and agriculture officers to understand. These explainability methods highlighted the exact diseased regions on cotton leaves, enabling farmers and field technicians to easily understand why the model made a particular prediction [14].

## VI. PROTOTYPE WEB APPLICATION

To illustrate the practical uses of the proposed framework, we developed a lightweight web-based application called CottonLeafVision, shown in Fig. 11. This prototype is a proof-of-concept to demonstrate the integration of automated cotton leaf disease detection into agricultural workflows. In Fig-11 we can see that the prototype features a simple drag-and-drop interface that enables users to upload images of cotton leaves. Upon submission, the system processes the input image using the trained DenseNet201 model and returns:

- The predicted cotton leaf disease and the confidence score made by the model for the associated leaf image
- A Grad-CAM heatmap overlay highlighting the leaf's infected or symptom areas [15].

Fig. 11. CottonLeafVision: A web app to classify Cotton Leaf Disease

## VII. Comparison and Discussion

In order to place the performance of the proposed CottonLeafVision framework with the existing research works, we conducted a comparative analysis with the three CNN architectures assessed in this study and also compared with previously established methods for Cotton Leaf Disease classification.

### A. *Evaluation of CNN Architectures*

Table III shows a comparative analysis of the performance metrics for DenseNet201, InceptionV3, VGG19. DenseNet201 exhibited the highest test accuracy. DenseNet201 outperformed InceptionV3(97%) and VGG19(93%) with the accuracy of 98%. Although the differences in performance were relatively moderate, they were consistently monitored across precision, recall, and F1-score. DenseNet201 exhibited a low rate of misclassification among all the diseases, indicating its dense connectivity effectively captures minor variations. These results answer the first research question by illustrating that CNN-based architecture can be optimized to achieve high accuracy and consistent performance in the classification of cotton leaf diseases under real-field variability.

TABLE III. Performance Comparison of Evaluated CNN Architectures

| Model Name | Classes | Precision | Recall | F1-Score |
|---|---|---|---|---|
| DenseNet201 | Bacterial Blight | 0.97 | 0.97 | 0.97 |
| | Curl Virus | 0.97 | 0.99 | 0.98 |
| | Healthy Leaf | 0.97 | 0.98 | 0.98 |
| | Heritage Growth Damage | 1.00 | 1.00 | 1.00 |
| | Leaf Hopper Jassids | 0.98 | 0.98 | 0.98 |
| | Leaf Redding | 0.99 | 0.95 | 0.97 |
| | Leaf Variegation | 0.99 | 1.00 | 1.00 |
| InceptionV3 | Bacterial Blight | 0.96 | 0.98 | 0.97 |
| | Curl Virus | 0.97 | 0.99 | 0.98 |
| | Healthy Leaf | 0.97 | 0.97 | 0.97 |
| | Heritage Growth Damage | 1.00 | 1.00 | 1.00 |
| | Leaf Hopper Jassids | 0.99 | 0.91 | 0.95 |
| | Leaf Redding | 0.92 | 0.95 | 0.94 |
| | Leaf Variegation | 0.98 | 0.99 | 0.99 |
| VGG19 | Bacterial Blight | 0.86 | 0.96 | 0.91 |
| | Curl Virus | 0.98 | 0.86 | 0.91 |
| | Healthy Leaf | 0.95 | 0.99 | 0.97 |
| | Heritage Growth Damage | 0.99 | 0.98 | 0.98 |
| | Leaf Hopper Jassids | 0.86 | 0.84 | 0.85 |
| | Leaf Redding | 0.89 | 0.88 | 0.88 |
| | Leaf Variegation | 0.99 | 0.99 | 0.99 |

### B. *Benchmarking Against Prior Work*

We have done a comparative analysis of CottonLeafVision and existing studies on cotton leaf disease classification and grading. As shown in Table IV, the DenseNet201 achieved an accuracy of 98%, surpassing several previous methods. For instance, Bishshash [1] reported an accuracy of 96.03% using InceptionV3, whereas Abudukelimu [2] achieved 93.3% using CM-YOLO. Herok, A. Kaur [3] achieved 95.50% utilizing the VGG16 framework. In contrast, CottonLeafVision demonstrated superior accuracy and minimized misclassification among cotton leaf diseases.

One significant difference in the current research is that many studies have primarily focused on predicted accuracy, avoiding the matter of robustness and interpretability. By applying adversarial robustness analysis and implementing Grad-CAM visualizations, CottonLeafVision advances this subject. This method answers the second research question. It demonstrates how CottonLeafVision outperforms existing CNN architectures for cotton leaf disease detection with the capability essential for agricultural implementation by combining high accuracy, precision, reliability, and stability.

TABLE IV. Comparison of MediVision with Previous Works on Cotton Leaf Disease Classification

| Author | Method | Accuracy (%) |
|---|---|---|
| Bishshash [1] | InceptionV3 | 96.03 |
| Abudukelimu [2] | CM-YOLO | 93.3 |
| A. Kaur [3] | VGG16 | 95.50 |
| Proposed Model | VGG19 | 93 |
| | InceptionV3 | 97 |
| | DenseNet201 | 98 |

### C. *Discussion of Limitations and Future Work*

This study had several limitations. The dataset was constrained in size and exhibited class imbalance, which may have influenced the residual misclassifications. Additionally, the evaluation relied on a single public dataset without external or multicenter validation, thereby limiting generalizability. Although adversarial perturbation tests confirmed the stability against noise, further investigation is required to assess the domain shifts across the imaging centers and devices. Future research should focus on enhancing dataset diversity, incorporating uncertainty quantification, and involving clinicians in usability testing.

## VIII. Conclusion

This study introduced CottonLeafVision, a deep learning tool that leverages the pretrained DenseNet201 model and helps to classify and detect the cotton leaf disease from a natural image. DenseNet201 achieved a test accuracy of 98% and was robust against the small changes in the images. Grad-CAM images and occlusion sensitivity tests were used to simplify the interpretation of the results. These results show that the decisions of the model are based on the important features of the leaf. A web application was also created to demonstrate how CottonLeafVision can be used as a tool to help agricultural experts and farmers make decisions. Compared to the prior work, this model gives both reliability and robustness to the end users.

## REFERENCES

[1] Bishshash, P., Nirob, A. S., Shikder, H., Sarower, A. H., Bhuiyan, T., &; Noori, S. R. H. (2024). A comprehensive cotton leaf disease dataset for enhanced detection and classification. Data in Brief, 57, 110913.

[2] Abudukelimu, H., Zhang, G., Abulizi, A. et al. Cotton leaf disease detection model focusing on small targets and comprehensive feature extraction. Sci Rep 15, 41125 (2025). https://doi.org/10.1038/s41598-025-24898-5

[3] A. Kaur, V. Kukreja, M. Kumar, A. Choudhary and R. Sharma, "A Fine-tuned Deep Learning-based VGG16 Model for Cotton Leaf Disease Classification," 2024 IEEE 9th International Conference for Convergence in Technology (I2CT), Pune, India, 2024, pp. 1-6, doi:10.1109/I2CT61223.2024.10544051.

[4] Herok, A., &; Ahmed, S. (2023, September). Cotton leaf disease identification using transfer learning. In 2023 International conference on information and communication technology for sustainable development (ICICT4SD) (pp. 158-162). IEEE

[5] A. Kaur, R. Sharma, S. Chattopadhyay and K. Joshi, &;Cotton Leaf Disease Classification Using Fine-Tuned VGG16 Deep Learning Model,&; 2024 2nd World Conference on Communication &; Computing (WCONF), RAIPUR, India, 2024, pp. 1-4, doi:10.1109/WCONF61366.2024.10692185.

[6] C. M. Devi, S. P. Vishva and M. M. Gopal, &;Cotton Leaf Disease Prediction and Diagnosis Using Deep Learning,&; 2024 International Conference on Advances in Computing, Communication and Applied Informatics (ACCAI), Chennai, India, 2024, pp. 1-7, doi:10.1109/ACCAI61061.2024.10602390.

[7] N. Walia, R. Sharma, M. Kumar, A. Choudhary and V. Jain, &;Optimized VGG16 Model for Advanced Classification of Cotton Leaf Diseases, &; 2024 4th International Conference on Intelligent Technologies (CONIT), Bangalore, India, 2024, pp. 1-4, doi:10.1109/CONIT61985.2024.1062705

[8] Mehmood, S., Memon, F., Nighat, A., Memon, F. A., &; Saba, E. (2023). Comparative Analysis of Feature Extraction Methods for Cotton Leaf Diseases Detection. VFAST Transactions on Software Engineering, 11(3), 81–90. https://doi.org/10.21015/vtse.v11i3.1626

[9] Bishshash, Prayma; Nirob, Md Asraful Sharker; Shikder, Md. Habibur; Sarower, Afjal (2024), “SAR-CLD-2024: A Comprehensive Dataset for Cotton Leaf Disease Detection”, Mendeley Data, V2, doi: 10.17632/b3jy2p6k8w.2

[10] Kibria, G. (2025). MediVision: An Explainable and Robust Deep Learning Framework for Knee Osteoarthritis Grading. 2025 IEEE International Conference on Data and Software Engineering (ICoDSE). https://doi.org/10.1109/ICoDSE68111.2025.11351659

[11] Hamdi, M., Alsharif, N., Alsharif, M. H., & Jahid, A. (2024). Performance evaluation of E-VGG19 model: Enhancing real-time skin cancer detection and classification. Heliyon, 10(10), e31488. https://doi.org/10.1016/j.heliyon.2024.e31488

[12] Ghosh, S. (2025). Healthy harvests: A comparative look at guava disease classification using InceptionV3. IEEE. https://doi.org/10.48550/arXiv.2602.10967

[13] Shikdar, O. F. (2025). Enhancing tea leaf disease recognition with attention mechanisms and Grad-CAM visualization. International Conference on Computing and Communication Networks (ICCCNet-2025). Springer. https://doi.org/10.48550/arXiv.2512.17987

[14] Valois, P., Niinuma, K., \& Fukui, K. (2023). Occlusion sensitivity analysis with augmentation subspace perturbation in deep feature space (arXiv:2311.15022).arXiv. https://doi.org/10.48550/arXiv.2311.15022

[15] Alam, B. M. S. et al. (2025). Explainable deep learning for hog plumleaf disease diagnosis: Leveraging the elegance of the depth of Xception. https://doi.org/10.1109/GECOST66002.2025.11324532